\documentclass[10pt,twocolumn,letterpaper]{article}

\usepackage{booktabs}  
\usepackage{colortbl}
\usepackage[pagenumbers]{cvpr} 
\usepackage{amsmath}

\usepackage{mathrsfs}
\usepackage[ruled,vlined]{algorithm2e}

\definecolor{commentcolor}{RGB}{110,154,155}   
\newcommand{\PyComment}[1]{\ttfamily\textcolor{commentcolor}{\# #1}}  
\newcommand{\PyCode}[1]{\ttfamily\textcolor{black}{#1}} 

\definecolor{cvprblue}{rgb}{0.21,0.49,0.74}
\usepackage[pagebackref,breaklinks,colorlinks,allcolors=cvprblue]{hyperref}

\newcommand{\myparagraph}[1]{\smallskip\noindent\textbf{#1}}
\usepackage[ruled,vlined]{algorithm2e}

\definecolor{commentcolor}{RGB}{110,154,155}   
\renewcommand{\PyComment}[1]{\ttfamily\textcolor{commentcolor}{\# #1}}  
\renewcommand{\PyCode}[1]{\ttfamily\textcolor{black}{#1}} 

\title{Gen-SIS: \underline{Gen}erative \underline{S}elf-augmentation \underline{I}mproves \underline{S}elf-supervised Learning}

\newcommand*\samethanks[1][\value{footnote}]{\footnotemark[#1]}

\author{Varun Belagali$^{1}$\thanks{Equal contribution. Correspondence to \href{mailto:vbelagali@cs.stonybrook.edu}{vbelagali@cs.stonybrook.edu}\\ 
\hangindent=1.8em webpage: \href{https://histodiffusion.github.io/docs/publications/gensis}{https://histodiffusion.github.io/docs/publications/gensis}} , Srikar Yellapragada$^{1}$\samethanks[1], Alexandros Graikos$^1$, Saarthak Kapse$^1$,  Zilinghan Li$^2$, \\ Tarak Nath Nandi$^{2,3}$, Ravi K Madduri$^{2,3}$,  Prateek Prasanna$^1$, Joel Saltz$^1$,  Dimitris Samaras$^1$ \\ \\
$^1$Stony Brook University \quad $^2$Argonne National Laboratory \quad
$^3$ University of Chicago\\ 
}

\begin{document}
\maketitle

\begin{abstract}

Self-supervised learning (SSL) methods have emerged as strong visual representation learners by training an image encoder to maximize similarity between features of different views of the same image. To perform this view-invariance task, current SSL algorithms rely on hand-crafted augmentations such as random cropping and color jittering to create multiple views of an image. Recently, generative diffusion models have been shown to improve SSL by providing a wider range of data augmentations. However, these diffusion models require pre-training on large-scale image-text datasets, which might not be available for many specialized domains like histopathology. In this work, we introduce Gen-SIS, a diffusion-based augmentation technique trained exclusively on unlabeled image data, eliminating any reliance on external sources of supervision such as text captions. We first train an initial SSL encoder on a dataset using only hand-crafted augmentations. We then train a diffusion model conditioned on embeddings from that SSL encoder. Following training, given an embedding of the source image, this diffusion model can synthesize its diverse views. We show that these `self-augmentations', i.e. generative augmentations based on the vanilla SSL encoder embeddings, facilitate the training of a stronger SSL encoder. Furthermore, based on the ability to interpolate between images in the encoder latent space, we introduce the novel pretext task of disentangling the two source images of an interpolated synthetic image. We validate Gen-SIS's effectiveness by demonstrating performance improvements across various downstream tasks in both natural images, which are generally object-centric, as well as digital histopathology images, which are typically context-based.

\end{abstract}    
\section{Introduction}

\begin{figure}[!t]
\centering
    \includegraphics[width=\linewidth]{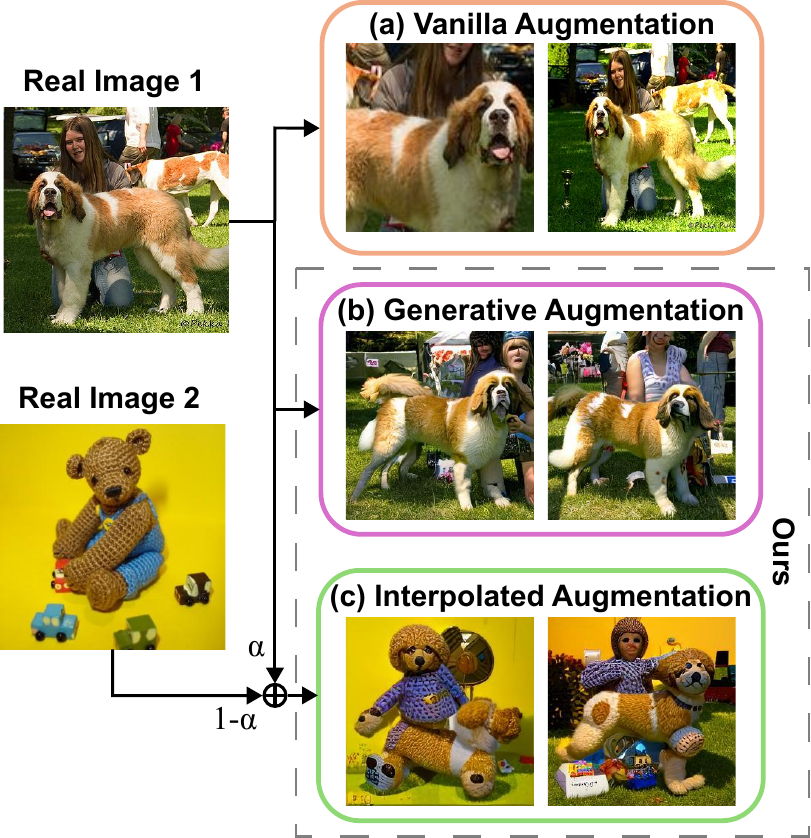} 
    \caption{(a) Vanilla augmentations used in SSL such as random cropping, color jittering. (b) Generative augmentations (ours) are conditioned on a single source image. (c) Interpolated augmentations (ours) conditioned on a pair of images. In the Gen-SIS framework, we use (b) for view augmentation, and (c) for the disentanglement pretext task, both in conjunction with (a).
    }
    
    \label{fig:teaser}
\end{figure}

In recent years, self-supervised learning (SSL)~\cite{dino, dinov2, byol, moco, jepa, simclr, mae, singh2023effectiveness} has emerged as a standard approach for learning robust visual representations that excel across various downstream tasks. By optimizing the model weights on pretext tasks, like self-prediction or view invariance, SSL enables models to learn discriminative features without requiring labeled data. Specifically, approaches such as DINO~\cite{dino}, BYOL~\cite{byol}, and SimCLR~\cite{simclr} have achieved notable success, producing high-quality features that transfer effectively to diverse downstream applications. This success stems from view-invariance tasks, which encourage models to learn high-level discriminative features from the image. Formulating view-invariant tasks relies heavily on hand-crafted augmentations, such as cropping and color jittering, to create multiple views of an image. \textit{Stronger augmentations typically lead to more robust features, as they increase the difficulty of the invariance task~\cite{byol}}.

On a parallel front, diffusion models have achieved impressive quality in image generation, driven by innovations in architecture~\cite{ldm, peebles2023scalable}, sampling methods~\cite{song2020denoising}, and conditioning techniques~\cite{ramesh2022hierarchical, ho2021classifier}. This success has led to an interest in using diffusion models, especially large foundation models like Stable Diffusion (SD), for data augmentation~\cite{trabucco2024effective, tian2024stablerep}. Given SSL's reliance on augmentations, diffusion models could significantly improve SSL by generating images with \textit{non-trivial} variations in background, shape, and position of objects, while preserving the original high-level semantics (Fig.~\ref{fig:teaser} (b)) 

Recent work by~\citet{tian2024stablerep} has investigated using synthetic data generated from Stable Diffusion (SD) as multiple views for SSL. However, employing SD as an augmenter in SSL has some drawbacks:
(1)~It is challenging to adapt SD in domains underrepresented in SD's training data, LAION-5B~\cite{schuhmann2022laion}. Since it is a general image foundation model, it is expected that it cannot generate high-quality images from specific domains such as histopathology. The low-quality images generated by SD cannot be used for SSL as they are highly inaccurate (see supplementary).
(2)~SD-scale foundation models are usually not available for other domains, outside natural images, and training them from scratch is a task beyond the scope of improving an SSL encoder.
(3)~Apart from synthesizing variations of an image, it is not straightforward to perform other kinds of augmentations by controlling the conditioning in text-to-image models. For instance, interpolating between two images would require using an LLM to first `interpolate' the two captions and then synthesize a new image.
(4)~As a text-conditioned model, SD is trained on paired image-text data, which can be seen as conflicting with the SSL principle of training on unlabeled data.

To avoid these issues, in this paper, we introduce Gen-SIS, a method to train a diffusion model on the same unlabeled data as an SSL model and use it as an effective augmenter for the SSL without any additional supervision, such as text or class labels. We adopt the term \textit{self-augmentation} to highlight the distinction between generative augmentations that rely on external supervision and our strictly self-supervised approach.

We begin by pre-training an SSL \textit{encoder} on real images from the pre-training dataset, using the original hand-crafted augmentations. Next, we train a latent diffusion model~\cite{ldm} (LDM), conditioned on image embeddings extracted from this initial SSL \textit{encoder}. Once trained, the LDM is then used to synthesize novel images for training a new enhanced SSL \textit{encoder}.

Gen-SIS expands the data augmentation using self-augmentations from the diffusion model, moving beyond traditional hand-crafted augmentations. In a view-invariant setting, a pair of real and synthetic images from our diffusion model can act as different views of the same image, strengthening the augmentation process (Fig.~\ref{fig:teaser} (b)). 

Furthermore, we utilize the generative model's capabilities and propose a novel pretext task that complements the base SSL task by focusing on disentangling shared concepts between pairs of images.
The trained LDM can interpolate between images by interpolating between the image embeddings provided as conditioning. The generated image semantically blends concepts from a given pair of real images (Fig.~\ref{fig:teaser} (c)). We then task the visual encoder with identifying features from the original pair of images used in generating the interpolated image. This additional pretext task (termed as \textit{disentanglement pretext task}) forces the model to learn and distinguish various object, texture, and shape-level features. Solving this task presents a greater challenge to the encoder, significantly enhancing its performance on downstream tasks. 

\noindent In summary, our contributions are: 
\begin{itemize}
    \item We introduce Gen-SIS, the first generative diffusion-enhanced SSL approach that requires only unlabeled data.
    \item We propose a novel disentanglement task, as an additional pretext task in self-augmentation enhanced SSL training.
    \item We extensively evaluate our method on ImageNet-1K and benchmark the Gen-SIS pretrained encoder across a range of downstream tasks such as classification, retrieval, copy detection, and video segmentation, achieving notable performance gains over vanilla SSL.
    \item Using Gen-SIS, we extend self-augmented SSL to histopathology images, a domain with no foundation generative models, demonstrating the effectiveness of our self-contained approach.
\end{itemize}

\begin{center}
 \begin{figure*}[!t]
 \centering
     \includegraphics[width=0.95\linewidth]{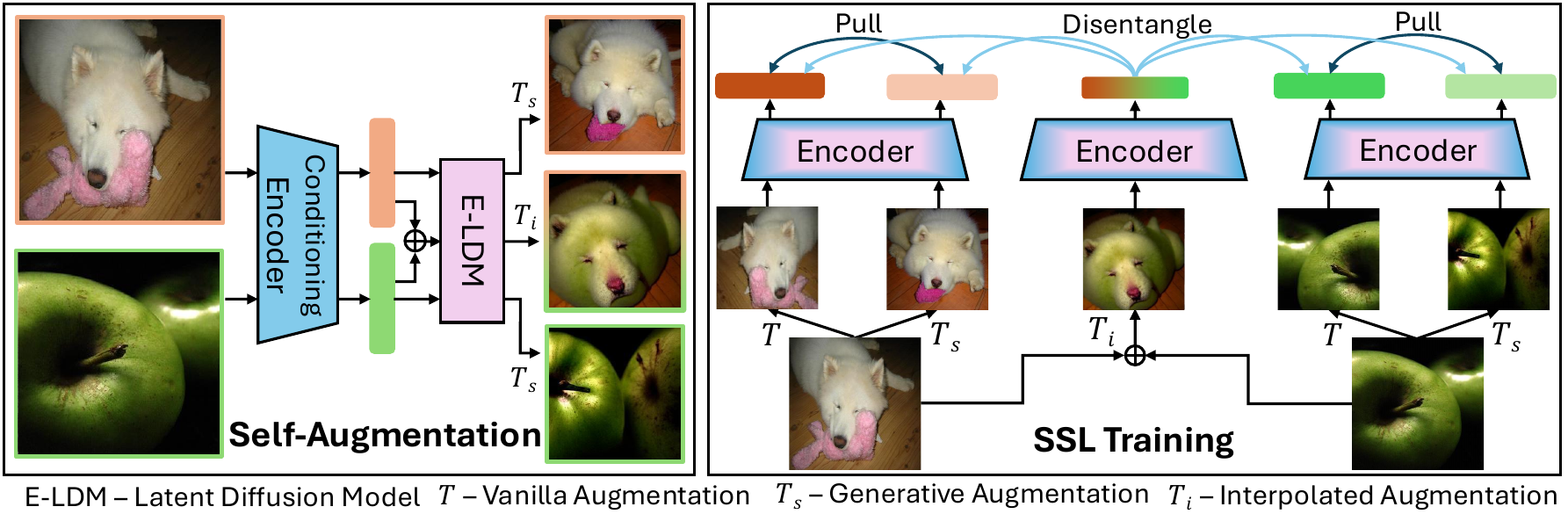} 
     \caption{Overview of the Gen-SIS-framework: It contains 2 key steps 1) Self-Augmentation using Embedding conditioned LDM (E-LDM), 2) SSL training with augmentations from E-LDM. $T$ represents vanilla augmentations, $T_{s}$ represents generative augmentation from single image, and $T_i$ represents interpolated augmentation from two images. Note that in conjunction with $T_{s}$ and $T_i$, we applied vanilla augmentation. \textit{Pull} represents the vanilla SSL pretext task, and \textit{Disentangle} represents our proposed pretext task with interpolated augmentation.
     }
     \label{fig:overview}
 \end{figure*}
 \end{center}

\section{Related work}
\label{sec:related_work}
 
\textbf{Self-supervised Learning:} Self-supervised learning~\cite{geiping2023cookbook} aims at learning generic representations from large-scale unlabeled data through a pretext task.  Pretext tasks can be mainly classified into self-prediction and view-invariance tasks. Self-prediction methods (MAE~\cite{he2022masked}, MaskFeat~\cite{wei2022masked}) involve masking parts of an image and then training the model to reconstruct the missing information based on the remaining context. View-invariant methods task the model to output similar features for two augmented views of the same image. This involves contrastive methods like SimCLR~\cite{chen2020simple}, MoCo~\cite{he2020momentum}, NNCLR~\cite{dwibedi2021little} and self-distillation methods like BYOL~\cite{grill2020bootstrap}, DINO~\cite{dino}, iBOT~\cite{zhou2021ibot}, and DINOv2~\cite{dinov2}. View-invariant methods typically rely on hand-crafted augmentations to derive multiple views of the same image for pretext tasks.

\noindent\textbf{Diffusion Models}: Diffusion models were first introduced in the seminal work of Ho \etal \cite{ho2020denoising}. Subsequent advancements included class-conditioning and guidance techniques for more controlled generation \cite{ho2022classifier, ramesh2022hierarchical}, and accelerated sampling techniques ~\cite{song2020denoising}. Latent diffusion models \cite{rombach2022high, peebles2023scalable, chen2024pixart} enable high-resolution image generation by performing the diffusion process in a smaller latent space.  In specialized domains such as histopathology, where labeled image-text data is limited, prior works have adopted image embedding-conditioned diffusion models~\cite{graikos2024learned, le2024infty} to overcome these constraints.

\noindent\textbf{Data augmentation with Diffusion models:} 
Recent research has utilized diffusion models for data augmentation, particularly in supervised settings \cite{trabucco2024effective, azizisynthetic, graikos2024learned, graikos2023conditional}. The studies most closely related to our research are Stable-rep \cite{tian2024stablerep} and SynCLR \cite{huangsynclr}. Stable-rep leverages captions from the CC-12M dataset to generate synthetic samples from Stable Diffusion~\cite{ldm} (SD), using them as multiple positive pairs in the SSL training. 

SynCLR, following a similar approach to Stable-Rep, uses ImageNet object categories to construct text prompts. 
However, SD-scale text-to-image models are usually unavailable beyond natural images.

Moreover, models trained on large-scale internet datasets, like LAION-5B, may accidentally contain examples from common benchmarks such as ImageNet. 
Previous works~\cite{carlini2023extracting, trabucco2024effective} have shown that pretrained diffusion models can leak training data, thus potentially inflating SSL performance.

\section{Preliminary}
\myparagraph{DINO:} In this study, we use DINO~\cite{dino} as our vanilla self-supervised learning (SSL) method. DINO (self-\textbf{di}stillation with \textbf{no} labels) is a teacher-student framework in which two augmented views of an image, \(I'\) and \(I''\), are processed separately by the student $g_{\theta_s}$ and teacher $g_{\phi_t}$ networks. The two augmented views are generated using standard augmentations, including cropping, color jittering, Gaussian blur, and solarization. Both teacher and student share the same architecture, with a backbone encoder and a projection head, and output a probability distributions $P$ over K dimensions.
\begin{gather}
    L_s = g_{\theta_s}(I'), \quad P_s^k = \frac{\exp(L_s^k/\tau_s)}{\sum_{j=1}^K \exp(L_s^j/\tau_s)}, \label{eq:student_prob} \\
    L_t = g_{\phi_t}(I''), \quad P_t^k = \frac{\exp((L_t^k - c^k)/\tau_t)}{\sum_{j=1}^K \exp((L_t^j - c^j)/\tau_t)},  \label{eq:teacher_prob}\\
    H(P_t,P_s) = -P_t \log(P_s),\quad \theta_s\leftarrow \text{Optimizer}(H,\theta_s) \label{eq:student_update} 
\end{gather}

The student's output (logits $L_s$)  is sharpened using a low-temperature $\tau_s$ softmax (Eq.~\ref{eq:student_prob}), while the teacher's output (logits $L_t$) undergoes centering with a moving average of the teacher outputs $c$ and softmax sharpening with $\tau_t$ to prevent collapse during training (Eq.~\ref{eq:teacher_prob}). The student network is optimized to match the teacher’s probability distribution using a cross-entropy loss $H$ (Eq.~\ref{eq:student_update}). The teacher network is updated as exponential moving average (EMA) of the student network's weights.

\myparagraph{Latent Diffusion Models:} Latent Diffusion Models (LDMs) \cite{ldm} synthesize images efficiently by learning to draw samples from a compressed image latent space instead of operating directly on pixels. This latent space is defined by a learned Variational Autoencoder (VAE), with a VAE encoder that maps images from pixels to latents, and a VAE decoder that maps the latent back to pixel space. Using the diffusion denoising objective \cite{ho2020denoising}, LDMs train a U-Net denoiser in the latent space. To control the generated images, the U-Net is usually conditioned on additional information about the images, such as class labels or text prompts. LDMs utilize a cross-attention mechanism between embeddings of the conditioning information and the U-Net features to guide the image synthesis, rendering the conditioning framework flexible to the choice of conditioning signals.

\section{Method}
\label{sec:method}
In this section, we introduce Gen-SIS (see Fig.~\ref{fig:overview}), a framework that leverages unlabeled data to train a diffusion model and subsequently enhances self-supervised learning (SSL) through novel self-augmentations using this learned diffusion model. First, in Sec.~\ref{subsec:eldm}, we describe the embedding-conditioned Latent Diffusion Model (E-LDM), which generates synthetic images based on the embeddings of source images. Then in Sec.~\ref{subsec:enhancingssl}, we detail how synthetic images (self-augmentations) generated by the E-LDM can be integrated into SSL to improve it. We focus on two types of self-augmentations: (1) Generative augmentations, where augmentations are created from a single source image, and (2) Interpolated augmentations, where an interpolated image is generated from two source images and used in training for a novel disentanglement pretext task.

\subsection{Embedding conditioned LDM}
\label{subsec:eldm}

We follow the LDM~\cite{ldm} framework for synthetic image generation, conditioning the LDM with the embedding extracted from an image, and refer to this setup as E-LDM (embedding-conditioned LDM). Following the approach of prior work~\cite{graikos2024learned}, we first train an image encoder on unlabeled real images using a standard SSL algorithm (DINO), and then use this encoder as the conditioning encoder to condition the diffusion model. This design allows our E-LDM to be trained in a fully self-supervised manner, without relying on any auxiliary information about the images. We term the synthetic images generated from E-LDM as self-augmentations. As conditioning, we choose the output of the DINO backbone, which is a $D$-dimensional vector $e$ (embedding). Once trained, we can then prompt the E-LDM by giving it an embedding of a real image $e$; it will synthesize a variation $I_s = \text{E-LDM}(z,e)$, where $z \sim \mathcal{N}(0,I)$ is an initial Gaussian noise used in sampling. We use the deterministic DDIM \cite{song2020denoising} sampling algorithm, which maps every $(z,e)$ pair to an image $I_s$.

\subsection{Enhancing SSL using self-augmentations}
\label{subsec:enhancingssl}

With real images as sources for E-LDM conditioning, we use two types of self-augmentations: 1) Generative Augmentations, 2) Interpolated Augmentations. 

\noindent \textbf{Generative Augmentations:} In the generative augmentation setting, a synthetic image is generated using a single real image as the source. This involves first extracting an embedding $e$ from the source image using the conditioning-encoder, and then guiding the image generation process with that embedding to create a synthetic image $I_s = \text{E-LDM}(z,e)$. As illustrated in Fig.~\ref{fig:teaser} (b), generative augmentations introduce novel variations in the shape, size, and position of objects, as well as changes in the background, while preserving the semantic content of the objects in the image. As shown in Fig.~\ref{fig:overview}, to integrate generative augmentations into SSL, we use the real image and a corresponding synthetic image as an input pair for the SSL pretext task. We also apply hand-crafted augmentations to both real and synthetic images. 

\noindent \textbf{Interpolated Augmentations:} An interesting property of diffusion models is their ability to generate an image that partially resembles each source image when conditioned on embeddings interpolated from two sources, as demonstrated in prior works~\cite{hudson2024soda, wang2023interpolating, graikos2024learned}. We leverage this property to produce an interpolated synthetic image from two real source images, which we use to perform a new pretext task during the SSL training. With embeddings $e_1$ and $e_2$ representing the two source images ($I_{1}$, $I_{2}$), and an interpolation ratio $\alpha$, we compute an interpolated embedding $e_{\text{int}}$ using spherical linear interpolation (SLERP)~\cite{wang2023interpolating} $e_{\text{int}} = \text{SLERP}(e_1, e_2, \alpha)$. We choose SLERP over linear interpolation since high-dimensional vectors are concentrated near the surface of the unit sphere. This interpolated embedding serves as the conditioning to generate the synthetic interpolated image, $I_{\text{int}} = \text{E-LDM}(z,e_{\text{int}})$. 

Since the interpolated image contains components of both source images, we propose a disentanglement task where the network learns to separate the distinct features of each source image used in the interpolation. Specifically, given two source images ($I_{1}$, $I_{2}$), an interpolating ratio ($\alpha$), and the interpolated synthetic image ($I_{\text{int}}$), we pass $I_{\text{int}}$ through the student network, to obtain the student probability $P_{\text{int}}$.
\begin{equation}
    L_{\text{int}} = g_{\theta_s}(I_{\text{int}}), \quad P_{\text{int}}^k = \frac{\exp(L_{\text{int}}^k/\tau_s)}{\sum_{j=1}^K \exp(L_{\text{int}}^j/\tau_s)}
    \label{eq:student_inter}
\end{equation}
To derive a target teacher output for the disentanglement task, we pass $I_{1}$, $I_{2}$ to the teacher network individually, and interpolate the teacher head output (logits $L_{\text{ent}}$) using $\alpha$:
\begin{equation}
    L_{\text{ent}} = \alpha g_{\phi_t}(I_{1}) + (1-\alpha) g_{\phi_t}(I_{2}).
    \label{eq:teacher_inter}
\end{equation}
This is then passed through the centering and sharpening operation to get the probability over the K dimensions
\begin{equation}
    P_{\text{ent}}^k = \frac{\exp((L_{\text{ent}}^k - c^k)/\tau_t)}{\sum_{j=1}^K \exp((L_{\text{ent}}^j - c^k)/\tau_t)}
    \label{eq:sharpen}
\end{equation}
Finally, we compute the disentanglement loss Eq.\ref{eq:disentangle} using the cross-entropy between the student and teacher predictions. 
\begin{equation}
    \mathcal{L}_{\text{disentangle}} = - P_{\text{ent}} \log (P_{\text{int}})
    \label{eq:disentangle}
\end{equation}

To optimize this loss, the student must implicitly disentangle components of the pair of source images within the interpolated image, leading us to call this a \textit{disentanglement pretext task}. 
This task is more challenging and can yield better representation learning compared to optimizing solely for single-source augmentations. With single-source images, the student only needs to extract features for a single dominant component to minimize the loss, whereas disentangling multiple components in an interpolated image can help the model learn more discriminative features. 

In Gen-SIS, we use both types of self-augmentations, generative augmentation with vanilla dino loss and interpolated augmentation with  $\mathcal{L}_{\text{disentangle}}$. We provide the pseudo code in the supplementary.

\section{Experiments: Natural Images}
\label{sec:results}

In this section, we apply the Gen-SIS framework to enhance SSL pre-training in the natural image domain. Our experiments below empirically demonstrate improvements in encoder pre-training using Gen-SIS compared to the vanilla DINO on diverse downstream tasks: classification, retrieval, copy detection, and video segmentation. We also provide evaluation on out-of-distribution data in the supplementary. Although we conduct experiments with DINO, our self-augmentation technique is a general method that can be readily extended to other SSL approaches. 

\subsection{Setup}

\noindent \textbf{Training:} Aligning closely with the experimental setup of DINO~\cite{dino}, we pre-train the models on the ImageNet-1K dataset~\cite{deng2009imagenet}. To begin, we reproduce the pre-training of ViT-S/16 model using the DINO framework (trained only on real images) on a 100 epoch setting with DINO's codebase. We use this model as the baseline and conditioning encoder for our E-LDM. For our enhanced SSL training, we improve DINO with the Gen-SIS framework and call the method  Gen-DINO. In Gen-DINO, we pre-train the ViT-S/16 model with generative and interpolated augmentations. Both DINO and Gen-DINO are trained for 100 epochs from scratch with a cosine annealing learning rate schedule with an initial value of $5 \times 10^{-4}$, a 10-epoch warmup period, and a linear scaling rule with respect to the batch size~\cite{chen2020big}. The weight decay also follows a cosine schedule, from 0.04 to 0.4. We use the AdamW optimizer with a batch size of 1024. We use generative and interpolation augmentation in Gen-SIS, in conjunction with the default handcrafted data augmentations of DINO, such as color jittering, cropping, flipping, Gaussian blur, solarization, and multi-crop. For both vanilla DINO and Gen-DINO, by default, we use 8 local crops; in ablations, we further show the performance without using local crops. For interpolated image generation, we use $\alpha = 0.5$. 

We train the LDM as an embedding conditioned model following  \cite{graikos2024learned}. The LDM configuration includes a VQ-f4 autoencoder that downsamples images from $256 \times 256 \times 3$ to $64 \times 64 \times 3$. For ImageNet experiments, we train the U-Net denoiser from scratch. We set the learning rate to $10^{-4}$ with a warmup period of 1000 steps. To generate images, we use DDIM sampling \cite{song2020denoising} with 50 steps and apply classifier-free guidance \cite{ho2021classifier}. We generate self-augmentations using E-LDM in an offline manner and read them from the disk during the Gen-DINO training. More details are provided in the supplementary.

\noindent \textbf{Evaluation:} We employ standard protocols used in DINO~\cite{dino}, such as the training-free k-nearest neighbor classifier ($k$-NN) and training a linear classifier (linear-probing) on frozen features. As highlighted in the DINO paper, linear probing is sensitive to hyperparameter variations, and hence we consider $k$-NN to be the preferred choice for evaluation given its robustness.

\begin{table*}[!t]
  \centering
  \begin{minipage}[t]{0.33\linewidth}
    \centering
    \caption{Top-1\% accuracy on \textbf{ImageNet-1K} validation set for ViT-S pre-trained through DINO and Gen-DINO and evaluated using $k$-NN and linear probing (LP) evaluation. $k$-NN is a training free evaluation.}
    \resizebox{1\linewidth}{!}{
\begin{tabular}{cc|cc}
    \toprule
    \textbf{Method}  & \textbf{Epochs}  &  \textbf{$k$-nn} & \textbf{LP} \\
    \midrule
     DINO    & 100   &   69.4   &   74.0   \\
     \rowcolor{blue!10} 
     Gen-DINO     & 100   &   \textbf{70.9}   &  \textbf{74.5}  \\ 
    \bottomrule
  \end{tabular}
    }
    \label{tab:im1k-class}
  \end{minipage}%
  \hspace{0.01\linewidth} 
  \begin{minipage}[t]{0.35\linewidth}
    \centering
    \caption{\textbf{Image retrieval.} We compared the mAP on the Oxford (ROx) and Paris (RPar) datasets using frozen features from ViT-S pre-trained with DINO and Gen-DINO on ImageNet-1K.}
    \resizebox{1\linewidth}{!}{
      \begin{tabular}{cc|cc|cc}
        \toprule
        \textbf{Method} & \textbf{Epochs} & \multicolumn{2}{c|}{\textbf{ROx}} & \multicolumn{2}{c}{\textbf{RPar}} \\
        & & M & H & M & H \\
        \midrule
        DINO & 100 & 30.7 & 10.8 & 55.6 & 26.1 \\
        \rowcolor{blue!10}
        Gen-DINO & 100 & \textbf{33.3} & \textbf{11.2} & \textbf{57.2} & \textbf{26.9} \\
        \bottomrule
      \end{tabular}
    }
    \label{tab:retrieval}
  \end{minipage}
\hspace{0.01\linewidth} 
  \begin{minipage}[t]{0.28\linewidth}
    \centering
    \caption{\textbf{Copy detection.} We report performance (mAP) using the Copydays ``strong" subset~\cite{douze2009evaluation}. We compare the features from ViT-S pre-trained with DINO and Gen-DINO.}
    \resizebox{1\linewidth}{!}{
      \begin{tabular}{ccc|c}
        \toprule
        \textbf{Method} &  \textbf{Epochs}   & \textbf{Dim} &  \textbf{mAP}  \\
        \midrule
         DINO   &   100    &  768  & 80.2     \\
         \rowcolor{blue!10} 
         Gen-DINO   &   100   & 768 & \textbf{82.5}     \\
        \bottomrule
      \end{tabular}
    }
    \label{tab:copy}
  \end{minipage}
\end{table*}

\subsection{Comparing with DINO on ImageNet-1K} 
In ~\cref{tab:im1k-class}, we compare the performance of ViT-S (patch size of 16) pre-trained using our Gen-DINO method against the vanilla DINO method with a 100-epoch schedule on the ImageNet-1K validation set. We observe that, compared to DINO, our method performs significantly better on $k$-NN evaluation, with an improvement of 1.5\% in Top-1$\%$ accuracy. The linear probing evaluation shows an improvement of 0.5\%. This evaluation indicates that Gen-DINO enhances representation learning through generative and interpolated augmentations, particularly by learning to solve the more challenging pretext task of disentangling two objects in the object-centric images found in ImageNet-1K. We further demonstrate the improvements of individual components in ablations (\cref{subsec:ablations}).

\subsection{Nearest neighbor retrieval}
Here, we investigate the effectiveness of Gen-DINO in enhancing performance compared to DINO on tasks that rely on nearest neighbor retrieval. Specifically, we evaluate its impact on image retrieval and copy detection tasks. We closely follow the settings described in DINO~\cite{dino}.
\\ 

\noindent \textbf{Image Retrieval:} We utilized the Revisited~\cite{radenovic2018revisiting} Oxford and Paris image retrieval datasets~\cite{philbin2008lost}. We used the corresponding Medium (M) and Hard (H) splits with query/database pairs and reported the Mean Average Precision (mAP). In \cref{tab:retrieval}, we compare the performance of ViT-S pre-trained with DINO and Gen-DINO on ImageNet-1K, using them as off-the-shelf frozen encoders for retrieval on these datasets. Following feature extraction, we apply $k$-NN for retrieval. We observe that Gen-DINO features outperform DINO features for this retrieval task by up to $2.6\%$ on the medium split and up to $0.8\%$ on the hard split across the two datasets.

\noindent \textbf{Copy Detection:} We use the ``strong" subset of the INRIA Copydays dataset~\cite{douze2009evaluation} and report the mean average precision (mAP). The task is to identify images that have been distorted by blur, insertions, print and scan, among other modifications, similar to the protocol in DINO. We perform this task using cosine similarity on the frozen features obtained from ViT-S pre-trained with DINO and Gen-DINO. We use the concatenation of the output [CLS] token and the GeM~\cite{radenovic2018fine} pooled output patch tokens, resulting in a 768-dimensional descriptor for ViT-S. In \cref{tab:copy}, we show that compared to vanilla DINO, our method substantially improves performance by $2.3\%$.

\subsection{Discovering the semantic layout of scenes}
Previously, DINO~\cite{dino} demonstrated the emerging properties of self-supervised ViTs, particularly their ability to explicitly represent scene layouts, with object segmentation visible in the self-attention modules of the last block. Here, we investigate how Gen-SIS' disentanglement pretext task, based on interpolated images, further enhances the model's capability for object segmentation without any supervision.

\noindent \textbf{Video Instance Segmentation:} In \cref{tab:video}, we evaluate the segmentation capabilities of self-supervised ViTs with Gen-DINO and compare them to vanilla DINO. Specifically, we used the DAVIS-2017 video instance segmentation benchmark~\cite{pont20172017}. Following the experimental protocol in DINO, we segment scenes using a nearest-neighbor approach between consecutive frames, utilizing the frozen features for the output patch tokens. We observe that our Gen-DINO pre-trained ViT-S performs significantly better for both the  mean region similarity ($\mathcal{J}_m$) and mean contour-based accuracy ($\mathcal{F}_m$) metrics,  demonstrating the effectiveness of the disentanglement task in enabling the model to more accurately understand object layout. We also compared it to Gen-DINO without the disentanglement task, i.e., DINO with only generative augmentation, and found that it performed worse than Gen-DINO. This is additional evidence that the disentanglement pretext task improves performance in understanding object details.

\begin{table}[!ht]
  \centering
  \caption{\textbf{DAVIS 2017 Video Object Segmentation.} We compared the performance of frozen features from ViT-S pre-trained with DINO and Gen-DINO on ImageNet-1K for the task of video instance tracking. Mean region similarity ($\mathcal{J}_m$) and mean contour-based accuracy ($\mathcal{F}_m$) metrics are reported. We use an image resolution of 480p.}
  \resizebox{1\columnwidth}{!}{
  \begin{tabular}{cc|ccc}
    \toprule
    \textbf{Method} &  \textbf{Epochs}   & \textbf{$(\mathcal{J} \& \mathcal{F})_m$} &  \textbf{$\mathcal{J}_m$} & \textbf{$\mathcal{F}_m$} \\
    \midrule
     DINO   &   100    &  61.45  &  59.67 & 63.23  \\
     \rowcolor{blue!10} 
     Gen-DINO w/o disent.   &    100   & 61.66 & 59.87 &  63.45  \\
     \rowcolor{blue!10} 
     Gen-DINO   &    100   & \textbf{62.07} & \textbf{60.52} &  \textbf{63.62}  \\
    \bottomrule
  \end{tabular}
  }
  \label{tab:video}
\end{table}

\begin{figure}[!t]
\centering
    \includegraphics[width=\linewidth]{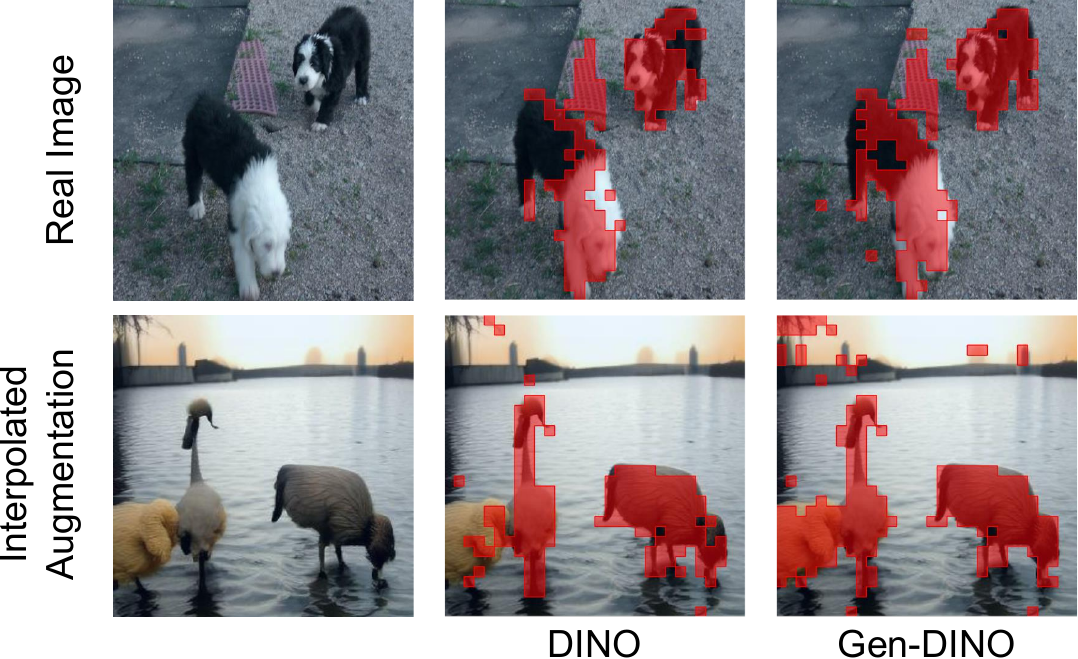} 
    \caption{ [CLS] token attention map of DINO and Gen-DINO averaged across all heads and overlayed on real and interpolated image. Gen-DINO's attention covers higher portion of object patches than DINO. 
    }
    \label{fig:attention}
\end{figure}

\noindent \textbf{Probing the self-attention map:} In Fig.~\ref{fig:attention}, we visualize the self-attention of the [CLS] token overlayed on a sample real image and on a sample interpolated image using pre-trained ViT-S using DINO and our Gen-DINO model. Consistently, for both real and generated images, Gen-DINO's attention map covers the object patches (16$\times$16 regions) more compared to DINO. This was also reflected in the significantly improved mean region similarity $\mathcal{J}_m$ in Tab.~\ref{tab:video}.

\subsection{Ablations}
\label{subsec:ablations}
Here, we study the effect of various components in Gen-DINO that are crucial for enhancing the performance of the encoder compared to vanilla DINO SSL. All ablations are conducted using ViT-S pre-trained for 100 epochs on ImageNet-1K and evaluated on its validation set. Top-1\% $k$-NN classifier accuracy is reported.

\noindent \textbf{Importance of Disentanglement pretext task:} In \cref{tab:ablation_disentanlge}, we investigate the effect of using only generative augmentation images without the proposed disentanglement pretext task (Gen-DINO wo/ disent.) and interpolated augmentation, comparing it to vanilla DINO and Gen-DINO. We observe that, by itself, generative augmentation provides a $0.5\%$ improvement compared to the larger $1.5\%$ improvement seen in Gen-DINO over vanilla DINO. This emphasizes that, beyond simple data augmentation, generative models can significantly enhance the SSL framework when used properly (in our case the interpolation augmentation and disentanglement pretext task), motivating future research.

\begin{table}[!ht]
  \centering
  \begin{minipage}[t]{0.50\linewidth}
    \centering
    \caption[Effect of disentanglement in DINO]{Effect of disentanglement (disent.) pretext task in Gen-DINO.}
    \resizebox{1\linewidth}{!}{
      \begin{tabular}{cc}
        \toprule
        \textbf{Method}   &  \textbf{$k$-NN} \\
        \midrule
         DINO    & 69.4     \\
         Gen-DINO wo/ disent.   & 69.9 (+0.5)     \\
         \rowcolor{blue!10} 
        Gen-DINO     & \textbf{70.9} (+1.5)   \\ 
        \bottomrule
      \end{tabular}
    }
    \label{tab:ablation_disentanlge}
  \end{minipage}%
  \hspace{0.05\linewidth} 
  \begin{minipage}[t]{0.40\linewidth}
    \centering
    \caption[Effect of interpolation ratio on Gen-DINO]{Effect of interpolation ratio $\alpha$ on Gen-DINO.}
    \resizebox{1\linewidth}{!}{
      \begin{tabular}{cc}
        \toprule
        \textbf{$\alpha$}   &  \textbf{$k$-NN} \\
        \midrule
         {0.2, 0.4, 0.6, 0.8}   & 70.0    \\
         {0.4, 0.6}   & 70.1    \\
         \rowcolor{blue!10} 
         {0.5}   & \textbf{70.9}     \\
        \bottomrule
      \end{tabular}
    }
    \label{tab:ablation_interpolate}
  \end{minipage}
\end{table}

\begin{figure*}[!ht]
\centering
    \includegraphics[width=\linewidth]{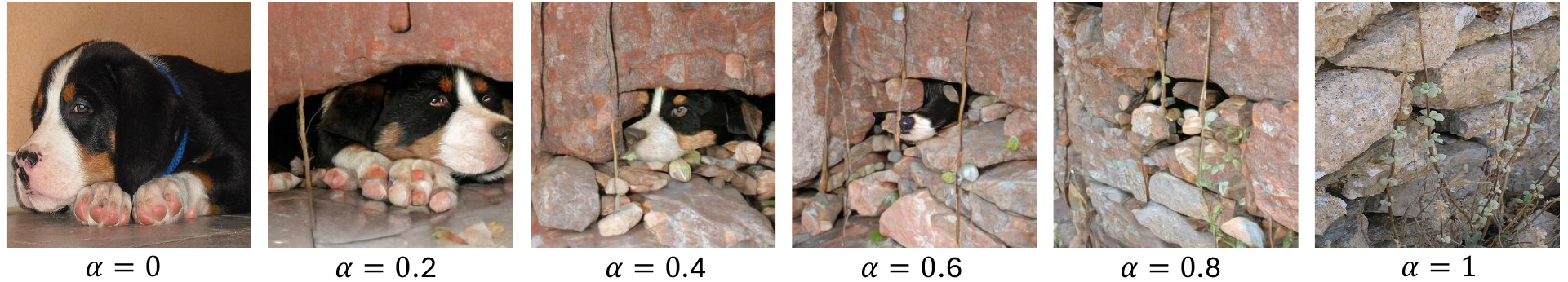} 
    \caption{Interpolated augmentations ($\alpha=\{0.2, 0.4, 0.6, 0.8\}$) generated from 2 real images ($\alpha$=0 and $\alpha$=1). An example of interpolation between dog and stone image from ImageNet dataset is illustrated. 
    }
    \label{fig:interpolate}
\end{figure*}

\noindent\textbf{Effect of Interpolation Ratio:} In Tab.~\ref{tab:ablation_interpolate}, we explore the effect of interpolation ratio ($\alpha$) in our framework. By default, we use $\alpha=0.5$ for interpolated image generation. However, other values or even randomly chosen values can be used as well. Therefore, we experiment with $\alpha=\{0.2, 0.4, 0.6, 0.8\}$ and $\alpha=\{0.4, 0.6\}$. We found that using values other than $\alpha=0.5$ reduces the model's performance.

To understand this drop, in Fig.~\ref{fig:interpolate}, we visualize the generated images with different $\alpha$ values. We observe that for values close to the boundaries (0.2 and 0.8), the interpolation is barely visible, with the image mostly gravitating toward the dominant side, making the pretext task noisy. The images synthesized with values 0.4 and 0.6 are very close to each other making it harder for the model to distinguish the exact $\alpha$ used in interpolation. Furthermore, this can also lead to noisy training if the interpolated images do not exactly reflect the interpolation ratio. We believe this is a limitation of the generative capabilities of the diffusion model for highly diverse datasets like ImageNet-1K~\cite{wang2023interpolating}. Hence, both intuitively and empirically, using $\alpha=0.5$ is the optimal solution as the SSL encoder only needs to understand that the interpolated image is a combination of two other images rather than finding the exact interpolation value. 

\noindent \textbf{Effect of teacher entanglement position:} In \cref{tab:ablation_position}, we experiment with the entanglement position of teacher outputs used in the disentanglement pretext task. By default, we entangle the teacher head logits (after the projection head) of two source images as per Eq.\ref{eq:teacher_inter}. We also explore performing the entanglement after the teacher backbone (before the projection head) and then passing the entangled embedding into the teacher head. ~\cref{tab:ablation_position} indicates that entangling before the projection head leads to a significant decrease in performance. This can be attributed to the low-dimensional teacher backbone output (384 in ViT-S), which allows less flexibility in feature entanglement within the low-dimensional space compared to the teacher projection head output, which is in much higher dimension (typically 65K).

\begin{table}[!ht]
  \centering
  \begin{minipage}[t]{0.45\linewidth}
    \centering
    \caption[Effect of teacher entanglement position]{Effect of teacher entanglement position.}
    \resizebox{1\linewidth}{!}{
      \begin{tabular}{cc}
        \toprule
        \textbf{Method}   &  \textbf{$k$-nn} \\
        \midrule
         Before proj. head   & 69.6     \\
         \rowcolor{blue!10} 
         After proj. head    & \textbf{70.9 }   \\ 
        \bottomrule
      \end{tabular}
      }
    \label{tab:ablation_position}
  \end{minipage}%
  \hspace{0.05\linewidth} 
  \begin{minipage}[t]{0.45\linewidth}
    \centering
    \caption[Comparison of DINO and Gen-DINO in case of only global crops.]{Comparison of DINO and Gen-DINO in case of only global crops.}
    \resizebox{1\linewidth}{!}{
      \begin{tabular}{cc}
        \toprule
        \textbf{Method}   &  \textbf{$k$-NN} \\
        \midrule
         DINO    & 58.6     \\
         Gen-DINO wo/ disent.   & 64.7     \\
         \rowcolor{blue!10} 
         Gen-DINO     & \textbf{67.4}   \\ 
        \bottomrule
      \end{tabular}
      }
    \label{tab:ablation_global_crops}
  \end{minipage}
\end{table}

\noindent \textbf{Training with Only Global Crops:} By default, the DINO method uses multiple local crops and has shown that they are necessary to achieve high performance; therefore, for a fair comparison, we include local crops in our framework as well. In Tab.~\ref{tab:ablation_global_crops}, we compare DINO and our method, Gen-DINO, without local crops. We observe that Gen-DINO performs significantly better, by $8.8\%$, compared to DINO. These findings could potentially help improve other SSL frameworks such as MoCov2~\cite{chen2020improved} and BYOL~\cite{grill2020bootstrap}, which do not benefit much or may even degrade when using local crops~\cite{dino}, but may improve with our generative and interpolated augmentations.
\\

\section{Experiments: Histopathology Imaging}
\label{sec:results_pathology} 

So far, we have evaluated Gen-DINO in the natural image domain, pre-training on the object-centric dataset ImageNet-1K. In this section, we explore its extension to histopathology, which is non-object-centric and instead involves a complex spatial layout of various tissue structures and nuclei types~\cite{chen2023general, kapse2024attention}. Given the lack of large-scale text-to-image foundation diffusion models in histopathology, self-augmentations using our Gen-SIS framework have a large potential to improve SSL in this domain.

\subsection{Setup}
\label{subsec:histology_setup}
\noindent \textbf{Dataset details:} We test our framework on two histopathology datasets: PANDA~\cite{bulten2022artificial} and BRIGHT~\cite{brancati2022bracs}. The PANDA dataset comprises approximately 10K prostate cancer whole-slide images (WSIs) with ISUP grading (6-class classification). The WSIs are sourced from two sites: Karolinska and Radboud. We use the slides from Karolinska for training and the slides from Radboud for evaluation. The BRIGHT dataset is a breast cancer dataset containing 703 WSIs, divided into 424 for training, 80 for validation, and 200 for testing. It features a 3-class classification (Non-cancerous, Pre-cancerous, and Cancerous) task. Due to the current inactivity of the BRIGHT challenge and the unavailability of test set labels, all results are reported using the validation set as the test set.

\noindent \textbf{Patch extraction and training:} WSIs are of gigapixel size and, therefore, need to be tiled into multiple crops to fit within hardware constraints. For the BRIGHT dataset, we use 10$\times$ magnification (1 micron per pixel), and for the PANDA dataset, we use 20$\times$ magnification (0.5 microns per pixel), extracting crops of size $256 \times 256$ pixels from each WSI. This yields 2M and 2.1M  crops for the train and test splits, respectively, for the PANDA dataset, and 1.2M and 0.2M million crops for the train and test splits, for the BRIGHT dataset. For both datasets, using the crops from the corresponding training set, we first pre-train a ViT-S from scratch with DINO, followed by training an E-LDM conditioned on this encoder. Finally, we pre-train a Gen-DINO using our Gen-SIS framework. We pre-train both DINO and Gen-DINO for 50 epochs on the PANDA dataset and 100 epochs on BRIGHT, using the same setting as ImageNet-1K. Following pre-training, we use the frozen encoders to extract embeddings for each crop in train-test set for both datasets. More details are provided in the supplementary.

\noindent \textbf{MIL setting:} Since we only have labels for each WSI, not individual crops, we treat a WSI as a bag of crops. We apply multiple instance learning (MIL)~\cite{ilse2018attention, thandiackal2022differentiable, kapse2024si, lu2021data, huang2024hard}, a method traditionally used in this context, to pool crop embeddings from each WSI and perform WSI-level prediction. For this task, we use ABMIL~\cite{ilse2018attention}. To ensure robustness, we conduct 5-fold cross-validation on PANDA and report mean performance on the test set. Since BRIGHT is a relatively small dataset in terms of the number of WSIs, we train MIL with 3 random seeds on the complete training set and report mean performance over the test set. The hyperparameter details used for MIL are provided in the supplementary.

\subsection{Results}
\label{subsec:histology_results}

As observed in \cref{tab:pathology}, MIL trained with features extracted from the Gen-DINO pre-trained encoder consistently outperforms those from the DINO pre-trained encoder across both datasets. In the PANDA dataset, our method improves performance by more than $3\%$ in balanced accuracy. In the BRIGHT dataset, we observe an improvement of $1.7\%$ in accuracy. It is important to note that the goal of this experiment is not to compare with the performance of foundational models on histopathology, but rather to enhance the DINO SSL method, which is a building block of all recent models in this field~\cite{chen2023general, filiot2024phikon, zimmermann2024virchow, lu2024visual}, with the potential to improve foundational models when Gen-DINO is scaled with larger datasets.
\begin{table}[!htbp]
\caption{Performance of DINO and Gen-DINO on PANDA (6-class classification) and BRIGHT (3-class classification) datasets. For PANDA, we report the mean over 5-fold cross-validation, and for BRIGHT, we report the mean over three seeds. }
\label{tab:pathology}
\begin{center}
\resizebox{\columnwidth}{!}{
\begin{tabular}{ccccccc}
\toprule
   \textbf{Method}      & \multicolumn{3}{c}{\textbf{PANDA}} & \multicolumn{3}{c}{\textbf{BRIGHT}} 
   \\
             & Acc.    & F1   & AUC    & Acc.  & F1     & AUC    \\  
\midrule

DINO          &    0.476  &    0.461   &    0.817  &    0.646   &    0.638  &    0.820   \\ 
\rowcolor{blue!10} 
   Gen-DINO             &     \textbf{0.508} &    \textbf{0.480}   &    \textbf{0.826}  &    \textbf{0.663}   &    \textbf{0.655}  &    \textbf{0.857}    \\ 

\bottomrule

\end{tabular}
}

\end{center}
\end{table}

\section{Conclusion}
\label{sec:conclusion}
We presented Gen-SIS, a self-augmentation technique to enhance self-supervised learning. Self-augmentations are generated from a diffusion model that does not rely on auxiliary information (text or class labels), making our approach a self-contained one. Our enhanced DINO (Gen-DINO) trained with Gen-SIS framework using generative augmentations, and interpolated augmentation along with the disentanglement pretext task outperforms the vanilla DINO in tasks such as image classification and nearest neighbor retrieval. More importantly, Gen-SIS pretraining enhances the self-supervised ViT's capability to explicitly represent semantic layout, as empirically proven through the video segmentation task. We showed that the disentanglement pretext task was the key contributor in enhancing this capability. We further extended our framework to non-object-centric histopathology images, showing consistent improvement across complex cancer grading tasks compared to DINO. Future work will explore novel approaches for flexible interpolation augmentation, including potential policies for selecting which image pairs to interpolate.

\section{Acknowledgements}
This research was partially supported by NCI awards
1R21CA258493-01A1, 5U24CA215109, UH3CA225021,
U24CA180924, NSF grants IIS-2123920, IIS-2212046, NIH 1R01CA297843-01, NIH NCI 1R21CA258493-01A1, Stony Brook Profund 2022 seed funding, and generous support from Bob Beals and Betsy Barton. This research used resources of the Argonne Leadership Computing Facility, a U.S. Department of Energy (DOE) Office of Science user facility at Argonne National Laboratory and is based on research supported by the U.S. DOE Office of Science-Advanced Scientific Computing Research Program, under Contract No. DE-AC02-06CH11357. We thank Jingwei Zhang for engaging in discussions on histopathology setup. We thank Prof. Beatrice Knudsen for annotating the descriptions of interpolated histopathology images presented in the supplementary.

{
    \small
    \bibliographystyle{ieeenat_fullname}
    \bibliography{main}
}

\maketitlesupplementarysingle




%



\noindent The supplementary is organized as follows:
\begin{itemize}
    \item Robustness evaluation (section \ref{supsec:robust})
    \item Comparison with pixel disentanglement (section \ref{supsec:pixel})
    \item Implementation details (section \ref{supsec:implementation})
\end{itemize}

\section{Robustness evaluation} \label{supsec:robust}
To evaluate the robustness of Gen-DINO, we benchmark its performance on three challenging datasets: ImageNet-A (Im-A) \cite{hendrycks2021nae}, ImageNet-R (Im-R) \cite{hendrycks2021many}, and ImageNet-Sketch (Im-S) \cite{wang2019learning}. These datasets test the model's resilience to out-of-distribution (OOD) variations. Im-A includes 7,500 adversarially filtered images across 200 classes of ImageNet. Im-R contains 30,000 images of renditions that are different from standard images from 200 classes of ImageNet. Sketch contains 50,000 black-and-white sketch images from all ImageNet classes. We directly evaluate the linear classifier trained on ImageNet-1K on these datasets. As shown in Tab.~\ref{tab:robustness}, Gen-DINO demonstrates improvements over the baseline on two robustness benchmarks. It achieves a substantial accuracy improvement on Im-R, increasing from 33.25 to 37.98, and a decent improvement on Sketch, rising from 61.93 to 62.3. These results suggest that the generative and interpolated augmentations in Gen-SIS enhance the model's ability to handle OOD images. Gen-DINO learns to encode more robust features, better capturing important characteristics of the images even under distribution shifts.

\begin{table}[!ht]
  \centering
  \caption{\textbf{Robustness.} We evaluate the linear classifier trained on ImageNet-1K. Gen-DINO shows notable improvements on ImageNet-R (Im-R) and Sketch (Im-S), indicating an enhanced ability to generalize to diverse image variations. }
  \resizebox{0.4\columnwidth}{!}{
  \begin{tabular}{cc|ccc}
    \toprule
    \textbf{Method} & \textbf{Epochs} & \textbf{Im-A}   & \textbf{Im-R} &   \textbf{Im-S}  \\
    \midrule
     DINO   &  100  & 9.24    &  33.25  & 61.93     \\
     \rowcolor{blue!10} 
     Gen-DINO   &  100 & 9.24   & \textbf{37.98} & \textbf{62.30}     \\
    \bottomrule
  \end{tabular}
  }
  \label{tab:robustness}
\end{table}

\section{Comparison with pixel disentanglement}\label{supsec:pixel}

An important question to address is whether a generative model is the optimal way to interpolate between images, or if simpler techniques, such as pixel-level interpolation, could achieve similar results. To investigate this, we perform an ablation comparing Gen-SIS's interpolated augmentations, performed in the conditioning space of E-LDM, against pixel-level interpolation of real images. In this regard, we train DINO with the same disentanglement pretext task (as proposed in Eq.7 of the main text) but replace embedding space interpolation with pixel-level interpolation. We refer to this model as ``\textit{DINO w/ pixel disent.}"

As shown in Tab.~\ref{tab:pixel_disentangle}, \textit{DINO w/ pixel disent.} significantly underperforms Gen-DINO by 3.0\% in terms of $k-$NN evaluation. This performance gap highlights the importance of interpolated augmentations in Gen-SIS, performed through E-LDM's conditioning space (embedding space). Interestingly, \textit{DINO w/ pixel disent.} improves linear probing accuracy over vanilla DINO by 0.59\% and achieves comparable performance to Gen-DINO in this metric. Improvement in linear probing of 0.4 \% over DINO has also been observed by a previous work~\cite{ren2022simple} that integrates interpolating real images in pixel space into DINO training. However, as noted by the authors of DINO, linear probing results are highly sensitive to hyperparameter tuning. Consequently, we prioritize $k$-NN evaluation as a more reliable metric. $k$-NN evaluation is training-free and provides a direct measure of the quality of learned representations, as its performance correlates with other downstream tasks like image retrieval and copy detection, which rely on nearest-neighbor comparisons in embedding space. The authors of DINOv2 \cite{dinov2} also emphasize using $k-$NN over linear probing to ablate key design choices.

\begin{figure}[!ht]
\centering
    \includegraphics[width=0.8\linewidth]{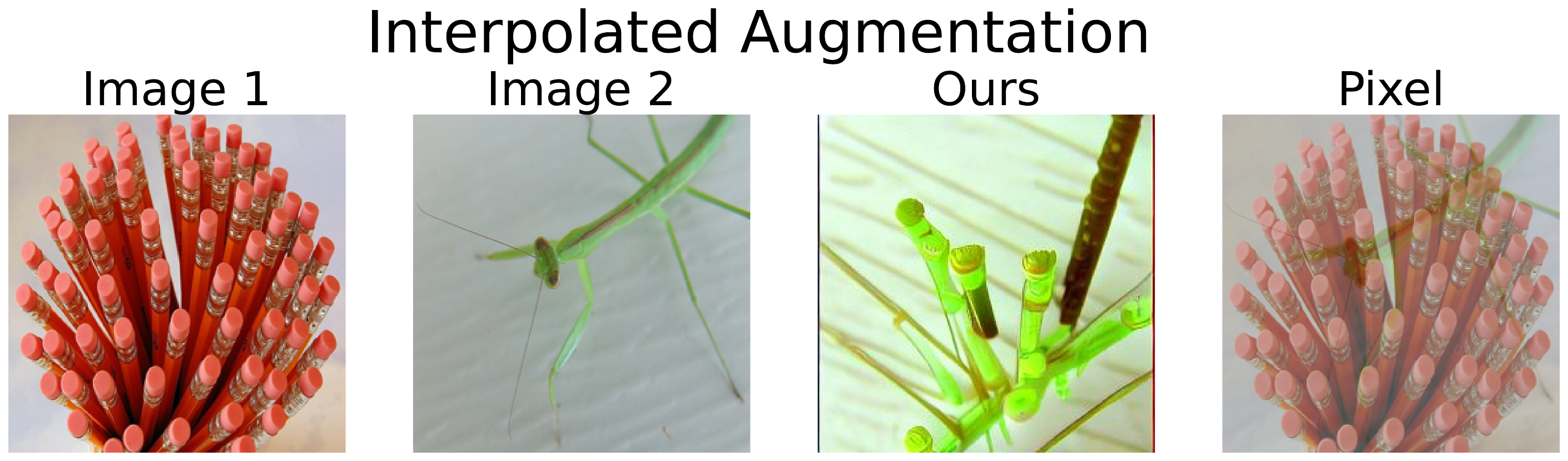} 
    \caption{ Interpolated augmentation using Gen-SIS framework (Ours) vs pixel-level interpolation. Image 1 and Image 2 are the source images used for interpolation ($\alpha=0.5$).
    }
    \label{fig:gensis_vs_pixel_sup}
\end{figure}

In Fig.~\ref{fig:gensis_vs_pixel_sup}, we visualize the interpolated augmentation under the Gen-SIS framework versus pixel-level interpolation. In Gen-SIS, the E-LDM blends the pencil (Image 1) and grasshopper (Image 2) to form a new object whose shape is similar to pencils, but color and texture follow the grasshopper. In pixel-level interpolation, the resulting textures and shapes are very different from the ones seen in the training images; (i) the edges are less prominent, due to the misaligned blending of the two images, and (ii) the textures are `unnatural' with mixtures of colors between the two images creating faded textures. Overall, we posit that the E-LDM tries to synthesize an image with objects formed from coherent blending of features from source image objects instead of the abruptly blended samples that pixel-level interpolation produces.

\begin{table}[!ht]
\centering
    \caption{Top-1\% accuracy on \textbf{ImageNet-1K} using DINO, DINO w/ pixel disent., and Gen-DINO. We report $k$-NN and linear probing (LP) evaluation.}
    \resizebox{0.4\linewidth}{!}{
      \begin{tabular}{cc|cc}
        \toprule
        \textbf{Method}   & \textbf{Epochs} & \textbf{$k$-NN} & \textbf{LP} \\
        \midrule
         DINO  & 100  & 69.4  &  73.97 \\
         DINO w/ pixel disent. & 100  & 67.9  & 74.56  \\
         \rowcolor{blue!10} 
        Gen-DINO  & 100   & 70.9 & 74.49 \\ 
        \bottomrule
      \end{tabular}
    }
    \label{tab:pixel_disentangle}
\end{table}

\begin{figure}[!ht]
\centering
    \includegraphics[width=\linewidth]{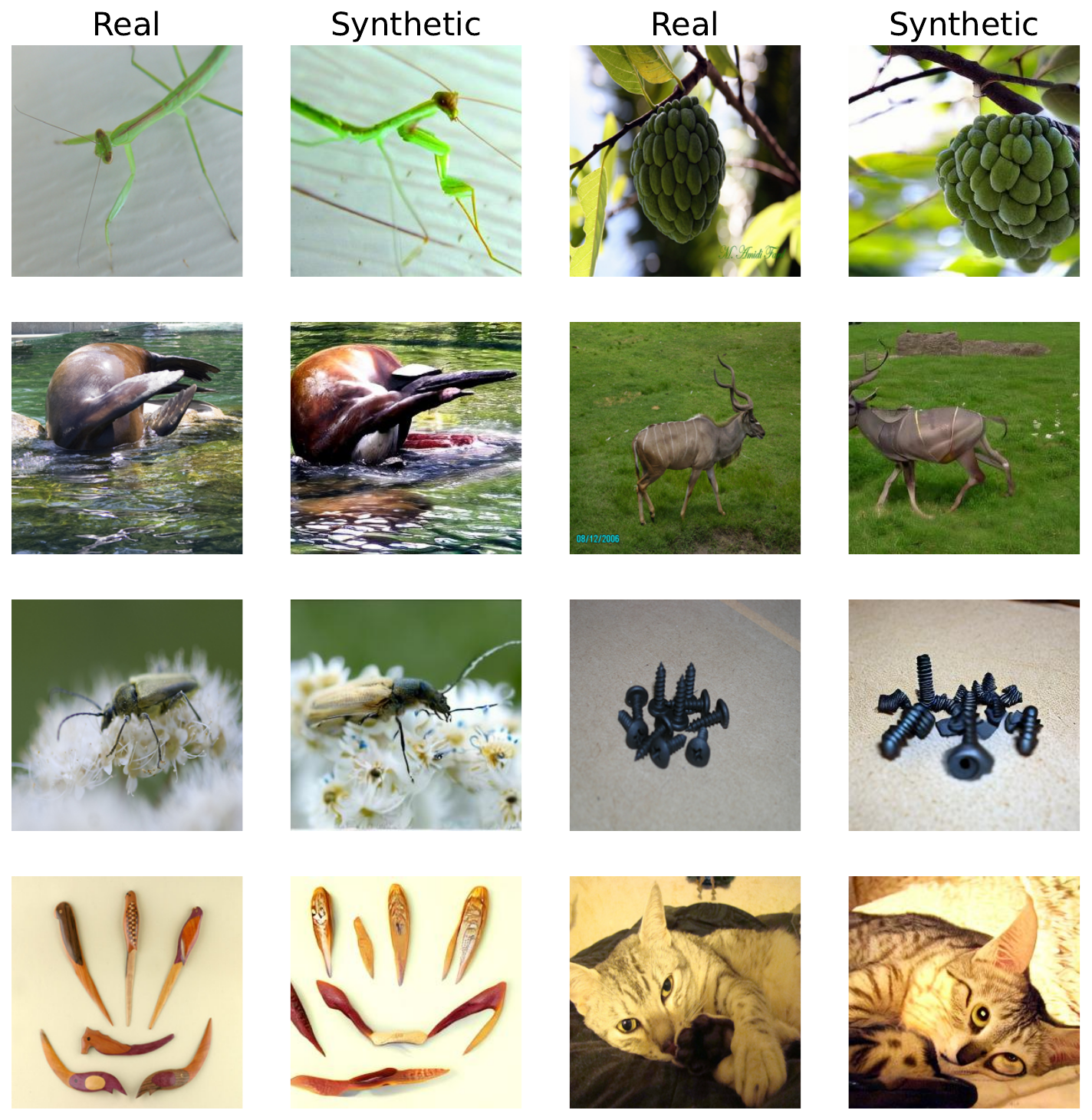} 
    \caption{ Generative Augmentation on ImageNet-1K using E-LDM by conditioning it on a single real image's embedding. Real: denotes the real image in the dataset, Synthetic: denotes the generative augmentation.
    }
    \label{fig:imagenet_gen_sup}
\end{figure}

\begin{figure}[!ht]
\centering
    \includegraphics[width=\linewidth]{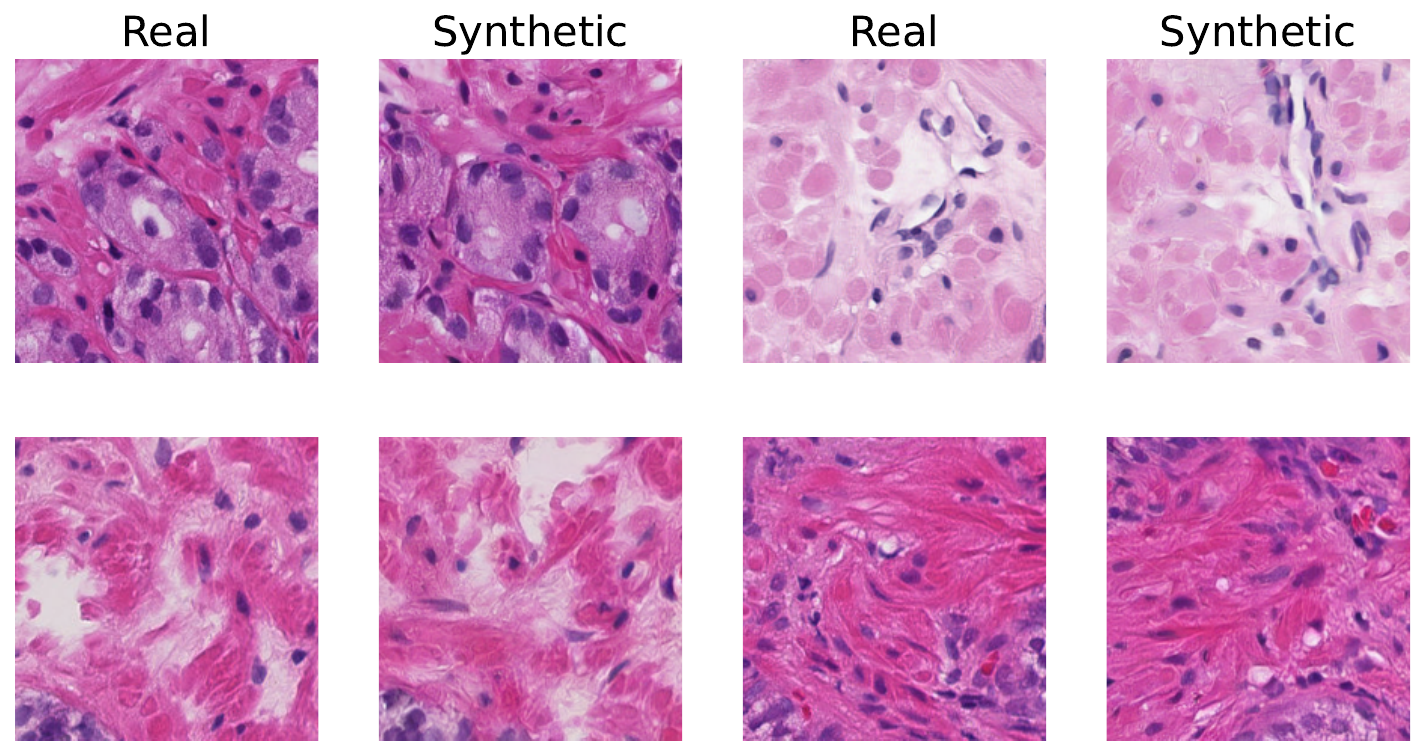} 
    \caption{ Generative Augmentation on PANDA using E-LDM by conditioning it on a single real image's embedding. Real: denotes the real image in the dataset, Synthetic: denotes the generative augmentation.
    }
    \label{fig:panda_gen_sup}
\end{figure}

\section{Implementation details}\label{supsec:implementation}
\subsection{Generation of self-augmentations}\label{supsec:augs}

\begin{center}
 \begin{figure*}[!ht]
 \centering
     \includegraphics[width=0.95\linewidth]{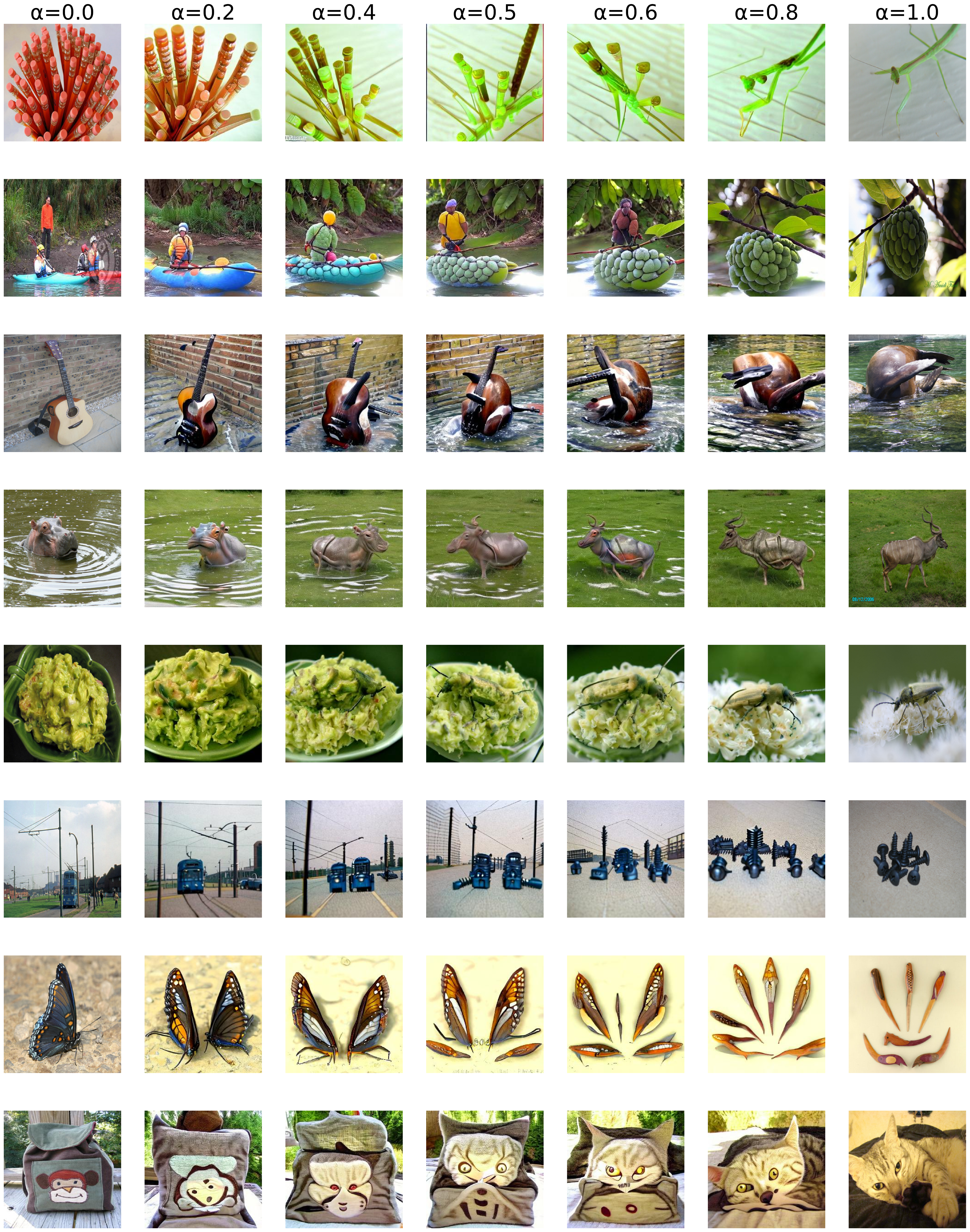} 
     \caption{Interpolated augmentations at various interpolating ratios on ImageNet-1K. $\alpha$=0.0 and $\alpha$=1.0 denote the two real images used as sources for interpolation. We interpolate the embeddings of the two source images, and then condition the E-LDM using the interpolated embedding to synthesize interpolated augmentations.}
     \label{fig:imagenet_interp_sup}
 \end{figure*}
 \end{center}

For ImageNet-1K, we synthesize four generative augmentations for each real image and save them to disk. We sample a random synthetic image out of four when training Gen-DINO. Fig.~\ref{fig:imagenet_gen_sup} shows sample synthetic image generation by E-LDM when using embedding from a single real image as conditioning. Synthetic images generated from E-LDM contain variations in orientation, object shape, and background compared to real images. In the case of interpolated augmentation, for each real primary image in the dataset, we pick a random secondary real image out of the whole dataset and perform the interpolated augmentation. We create a single interpolated augmentation for each primary image and interpolation ratio ($\alpha$), and then read the interpolated augmentation from the disk when training. Fig.~\ref{fig:imagenet_interp_sup} presents the interpolated augmentation with various $\alpha$ values. We use $\alpha$=0.0 and $\alpha$=1.0 to represent the two real images used as sources for interpolated augmentation. Interpolated augmentations blend the shape, texture, and color of the objects visible in the two source images to form new, blended objects. As seen in Fig.~\ref{fig:imagenet_interp_sup}, a key observation is that for $\alpha$=0.2 and $\alpha$=0.8 interpolated images are very similar to the closest source image and contain negligible components from the other end. Following the ablation in Tab.6 (in main text), we use $\alpha$=0.5 in the training of Gen-DINO. For both generative and interpolated augmentation, we use classifier-free guidance of 6 with 50 DDIM steps.

\begin{figure}[!ht]
\centering
    \includegraphics[width=\linewidth]{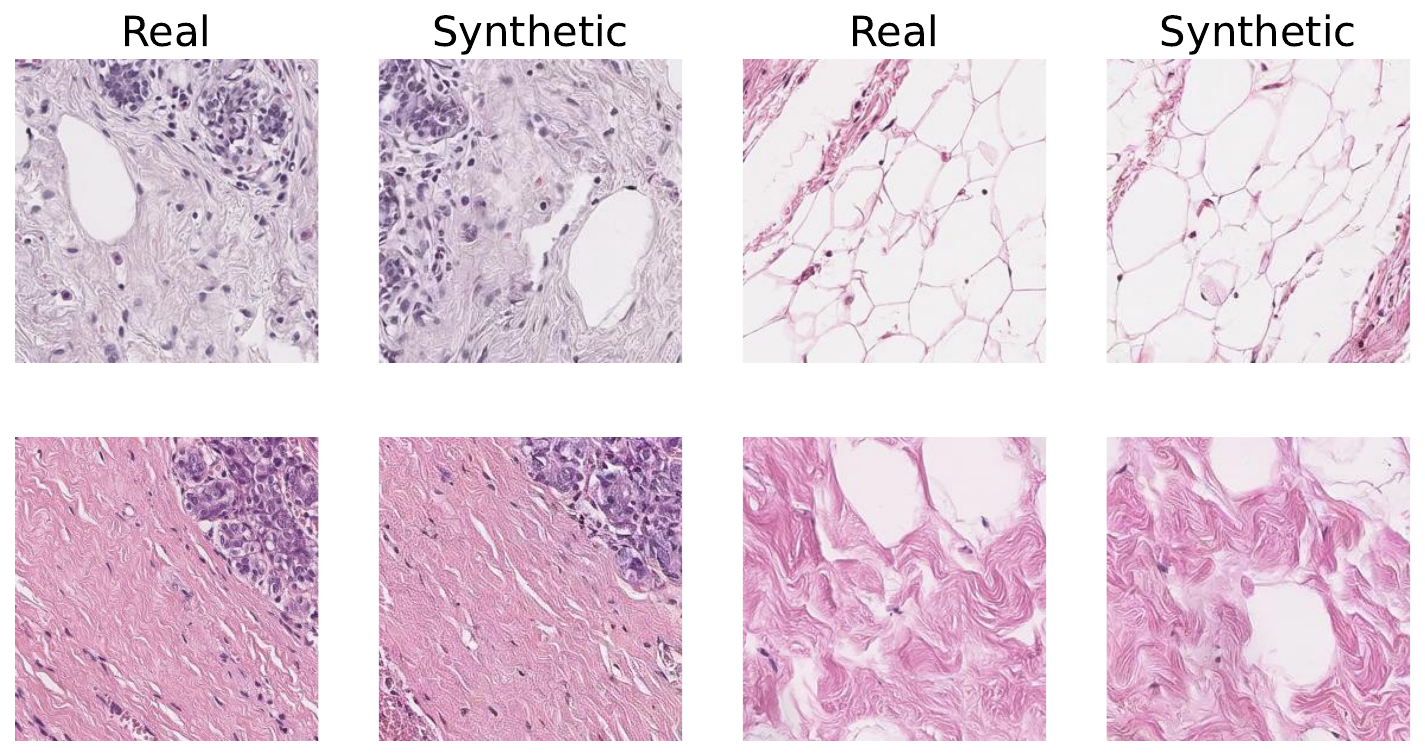} 
    \caption{ Generative Augmentation on BRIGHT using E-LDM by conditioning it on single real image's embedding. Real: denotes the real image in the dataset, Synthetic: denotes the generative augmentation.
    }
    \label{fig:bright_gen_sup}
\end{figure}

\begin{figure}[!ht]
\centering
    \includegraphics[width=\linewidth]{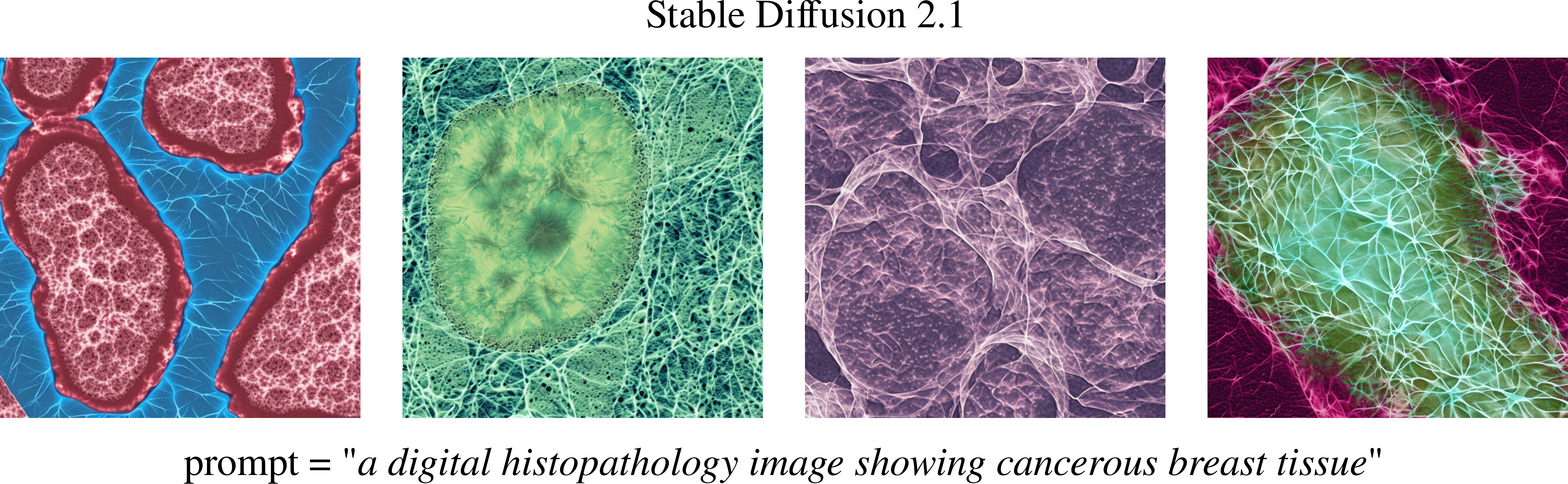} 
    \caption{ Breast cancer synthetic histopathology image generation using Stable Diffusion with text prompt as conditioning. The images generated do not resemble real breast cancer images that are found in typical datasets like BRIGHT (Fig~\ref{fig:bright_gen_sup}). 
    }
    \label{fig:sd_histo}
\end{figure}

\begin{center}
 \begin{figure*}[!ht]
 \centering
     \includegraphics[width=1\linewidth]{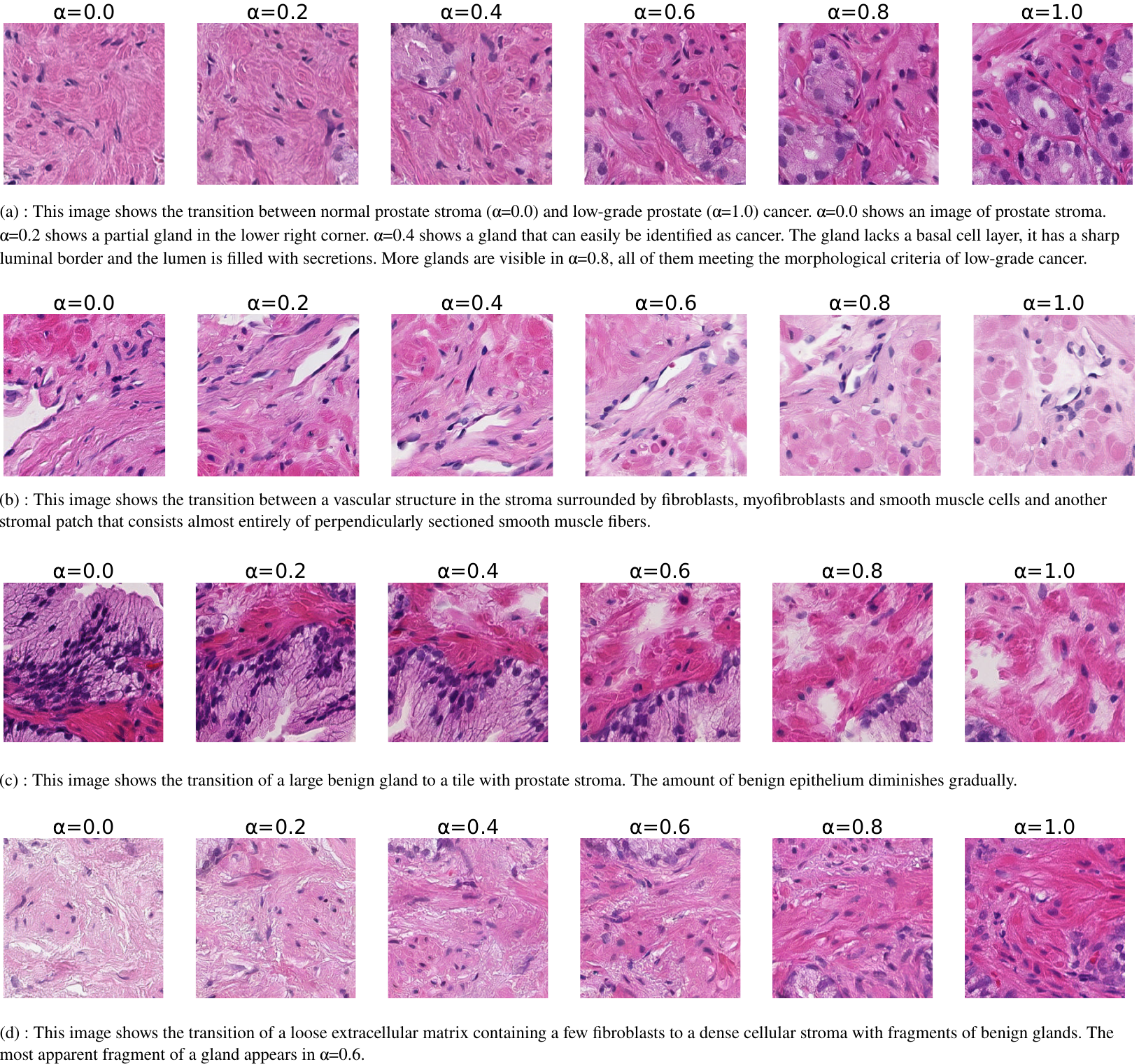} 
     \caption{Interpolated augmentations at various interpolating ratios on PANDA. $\alpha$=0.0 and $\alpha$=1.0 denote the two real images used as sources for interpolation. We interpolate the embeddings of the two source images, and then condition the E-LDM using the interpolated embedding to synthesize interpolated augmentations. The captions below each row represent the description of interpolation annotated by a pathologist.
     }
     \label{fig:panda_interp_sup}
 \end{figure*}
 \end{center}

\begin{center}
 \begin{figure*}[t]
 \centering
     \includegraphics[width=1\linewidth]{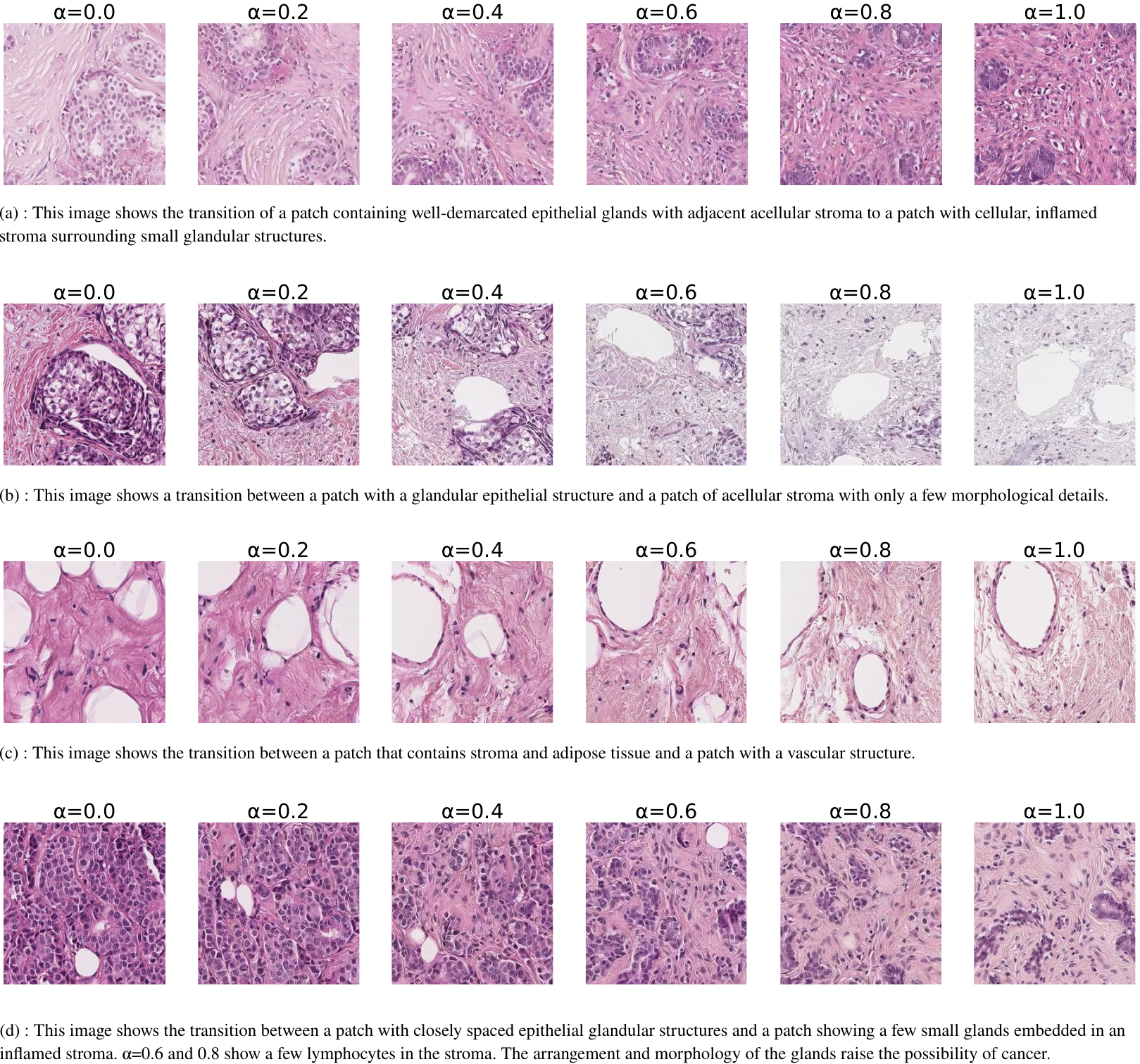} 
     \caption{Interpolated augmentations at various interpolating ratios on BRIGHT. $\alpha$=0.0 and $\alpha$=1.0 denote the two real images used as sources for interpolation. We interpolate the embeddings of the two source images, and then condition the E-LDM using the interpolated embedding to synthesize interpolated augmentations. The captions below each row represent the description of interpolation annotated by a pathologist.
     }
     \label{fig:bright_interp_sup}
 \end{figure*}
 \end{center}

In the case of histopathology (PANDA and BRIGHT datasets), we follow a similar setup as ImageNet-1K, and synthesize one generative augmentation and one interpolated augmentation for each image. Fig.~\ref{fig:panda_gen_sup} and Fig.~\ref{fig:bright_gen_sup} present generative augmentations, i.e., sample synthetic images generated using real images as source. In generative augmentation, the synthetic image varies in terms of the position and orientation of cells and tissue compared to the real source image. In the case of interpolation augmentation, for each real primary image (crop) in the dataset, we pick a random secondary real image (crop) from a different whole slide image and perform the interpolated augmentation. We create a single interpolated image for each primary image and given interpolation ratio ($\alpha$) and read the interpolated image from the disk when training. Fig.~\ref{fig:panda_interp_sup} and Fig.~\ref{fig:bright_interp_sup} showcase the interpolated augmentations in PANDA and BRIGHT datasets, respectively. Unlike ImageNet, we observe that even $\alpha$=0.2 and $\alpha$=0.8 interpolated images contain some components from lower-weighted source images. Following this observation, we sample a random alpha from \{0.2, 0.4, 0.6, 0.8\} for the interpolated augmentation during the training of Gen-DINO. For both generative and interpolated augmentation, we use a guidance weight of 1.75 with 50 DDIM steps following the recent works ~\cite{graikos2024learned, yellapragada2024pathldm} on diffusion models in histopathology. We also present sample synthetic breast cancer images generated from Stable diffusion with text prompts in Fig~\ref{fig:sd_histo}. The images are highly inaccurate to be used in training. This reinforces our key design choice of using E-LDM for augmentations.

\subsection{Pseudo code for disentanglement pretext task}\label{supsec:code}
Algorithm~\ref{algo:pseudocode_v2} presents the pseudo-code for the disentanglement pretext task. We only use global crops for this pretext task.

\begin{algorithm}[ht]
\caption{PyTorch-style pseudo-code for disentanglement pretext task}
\footnotesize
\SetAlgoLined
    \PyComment{Input image: img\_1}\\
    \PyComment{gs, gt: student and teacher networks} \\
    \PyComment{tps, tpt: student and teacher temperatures} \\
    \PyComment{c: center} \\
    \PyComment{alpha: interpolation ratio}\\
    
    \PyCode{for img\_1 in loader}\\
    \Indp

    \PyComment{Read secondary source image} \\
    \PyCode{img\_2 = ReadImage(secondary(img\_1)) }

    \PyComment{Read interpolated image of primary and secondary source image} \\
    \PyCode{img\_int = ReadInterpImage(img\_1, img\_2, alpha) }\\

    \PyComment{Apply vanilla dino augmentation to form a view of interpolation} \\
    \PyCode{img\_int\_view = vanilla\_augment(img\_int)} \\

    \PyComment{Apply vanilla dino augmentation to form a view of primary} \\
    \PyCode{img\_1\_view = vanilla\_augment(img\_1)} \\

    \PyComment{Apply vanilla dino augmentation to form a view of secondary} \\
    \PyCode{img\_2\_view = vanilla\_augment(img\_2)} \\

    \PyComment{Get student output for interpolated image and teacher output for image 1 and image 2} \\
    \PyCode{stud\_int = gs(img\_int\_view)} \\
    \PyCode{teach\_1 = gt(img\_1\_view).detach()} \\
    \PyCode{teach\_2 = gt(img\_2\_view).detach()} \\

    \PyComment{Student sharpening} \\
    \PyCode{stud\_int = softmax(stud\_int / tps, dim=1)} \\

    \PyComment{Entanglement of teacher output} \\
    \PyCode{teach\_ent = alpha * teach\_1 + (1-alpha) * teach\_2 } \\
    \PyComment{Teacher sharpening and centering} \\
    \PyCode{teach\_ent = softmax((teach\_ent - c) / tpt, dim=1)} \\

    \PyComment{Compute disentanglement loss} \\
    \PyCode{disentanglement\_loss = - (teach\_ent * log(stud\_int)).sum(dim=1).mean()} \\
\Indm
\label{algo:pseudocode_v2}
\end{algorithm}

\subsection{Evaluation details}\label{supsec:eval}
\noindent \textbf{ImageNet:} 
We employ standard protocols as used in DINO~\cite{dino}, such as the training-free k-nearest neighbor classifier ($k$-NN) and the learning of a linear classifier, both applied to frozen features. For $k$-NN evaluation, we extract the features from the training data using the frozen pre-trained encoder. Next, the $k$-NN classifier compares the features of an image to the $k$ nearest stored features and assigns a label. We explore various numbers of nearest neighbors and determine that 10-NN or 20-NN consistently yields the best results. In linear evaluation, random resize cropping and horizontal flip augmentation are applied during training, and test performance is reported on a central crop. We follow the same hyperparameter setup as DINO~\cite{dino}. We perform a learning rate hyperparameter search to find the optimal choice. As highlighted in the DINO paper, linear probing is sensitive to hyperparameter variations, and we similarly observe a substantial variance in accuracy across learning rate. Therefore, in our study, we consider $k$-NN as a preferable choice for evaluation, given its robustness to challenges like hyperparameter tuning. 

\noindent \textbf{Histopathology:} 
We employ multiple instance learning (MIL) to aggregate the frozen features of crops from a whole slide image, followed by a linear classifier applied to the pooled features. For our MIL framework, we utilize ABMIL~\cite{ilse2018attention}. The model is trained for 50 epochs using the AdamW optimizer with a learning rate of  $0.0001$ and a weight decay of $0.01$. Given that whole slide images can contain varying numbers of crops, we use a batch size of 1 and accumulate gradients over 8 steps, achieving an effective batch size of 8.

\end{document}